\documentclass[10pt, a4paper]{article}

\usepackage{fullpage}
\usepackage[utf8]{inputenc}
\usepackage{kotex}
\usepackage{kotex-logo}
\usepackage{verbatim}
\usepackage{fixltx2e}
\usepackage{booktabs}
\usepackage{multirow}
\usepackage{multicol}
\usepackage{graphicx}
\usepackage{amsmath}
\usepackage{amssymb}
\usepackage{xcolor}
\usepackage{colortbl}
\usepackage{arydshln}
\usepackage{tablefootnote}
\usepackage[most]{tcolorbox}
\usepackage{wrapfig,boxedminipage,caption}
\usepackage[utf8]{inputenc}
\usepackage{array}
\usepackage{enumitem}

\definecolor{Gray}{gray}{0.95}
\definecolor{commentcolor}{rgb}{0.1, 0.1, 0.9}
\newcolumntype{a}{>{\columncolor{Gray}}c}
\newcolumntype{b}{>{\columncolor{Gray}}l}
\newcolumntype{x}[1]{>{\raggedright\let\newline\\\arraybackslash\hspace{0pt}}m{#1}}
\newcolumntype{y}[1]{>{\raggedright\let\newline\\\arraybackslash\hspace{0pt}\columncolor{Gray}}m{#1}}

\newcommand{\astfootnote}[1]{
    \let\oldthefootnote=\thefootnote
    \setcounter{footnote}{1}
    \renewcommand{\thefootnote}{\fnsymbol{footnote}}
    \footnotetext{#1}
    \let\thefootnote=\oldthefootnote
}

\usepackage{lrec-coling2024} 

\title{BlendX: Complex Multi-Intent Detection with Blended Patterns}

\name{Yejin Yoon$^\dagger$, Jungyeon Lee$^\ddagger$, Kangsan Kim$^\ddagger$, Chanhee Park$^{\mathparagraph}$, Taeuk Kim$^{\dagger\ddagger*}$} 

\address{$^\dagger$Dept. of Al Application \& $^\ddagger$Dept. of Artificial Intelligence, Hanyang University \\
         $^{\mathparagraph}$Hyundai Motor Company\\
         Seoul, Republic of Korea \\
         \{stillwithyou, jungyune, kimtaeuk\}@hanyang.ac.kr\\}

\abstract{
Task-oriented dialogue (TOD) systems are commonly designed with the presumption that each utterance represents a single intent. 
However, this assumption may not accurately reflect real-world situations, where users frequently express multiple intents within a single utterance.
While there is an emerging interest in multi-intent detection (MID), existing in-domain datasets such as MixATIS and MixSNIPS have limitations in their formulation. 
To address these issues, we present BlendX, a suite of refined datasets featuring more diverse patterns than their predecessors, elevating both its complexity and diversity. 
For dataset construction, we utilize both rule-based heuristics as well as a generative tool---OpenAI's ChatGPT---which is augmented with a similarity-driven strategy for utterance selection.
To ensure the quality of the proposed datasets, we also introduce three novel metrics that assess the statistical properties of an utterance related to word count, conjunction use, and pronoun usage.
Extensive experiments on BlendX reveal that state-of-the-art MID models struggle with the challenges posed by the new datasets, highlighting the need to reexamine the current state of the MID field.
The dataset is available at \url{https://github.com/HYU-NLP/BlendX}.
 \\ \newline \Keywords{Multi-Intent Detection, Task-Oriented Dialogue, Spoken Language Understanding} }

\begin{document}

\maketitleabstract

\astfootnote{Corresponding author.}

\section{Introduction}
\label{sec:introduction}

The successful implementation of task-oriented dialogue (TOD) systems begins with the precise recognition of user intents.
By accurately discerning the queries embedded in user inputs and routing them to the relevant components, the systems can adeptly respond, thereby effectively fulfilling user requests. 
Generally, these systems are constructed on the assumption that each user utterance is exclusively linked to a single intent, which often diverges from practical scenarios.

Contrary to the conventional setting, the task of \textbf{Multi-Intent Detection (MID)} presents a more nuanced and comprehensive challenge for TOD systems, permitting users to express multiple intentions simultaneously.
The problems posed by MID are not only more demanding but also more realistic---for reference, \citet{gangadharaiah-narayanaswamy-2019-joint} reported that over half of the total instances (52\%) from Amazon's in-house dialogue dataset contain multiple intents, underscoring the practical significance of the task.

\begin{figure}
    \begin{center}
        \includegraphics[scale=0.45]{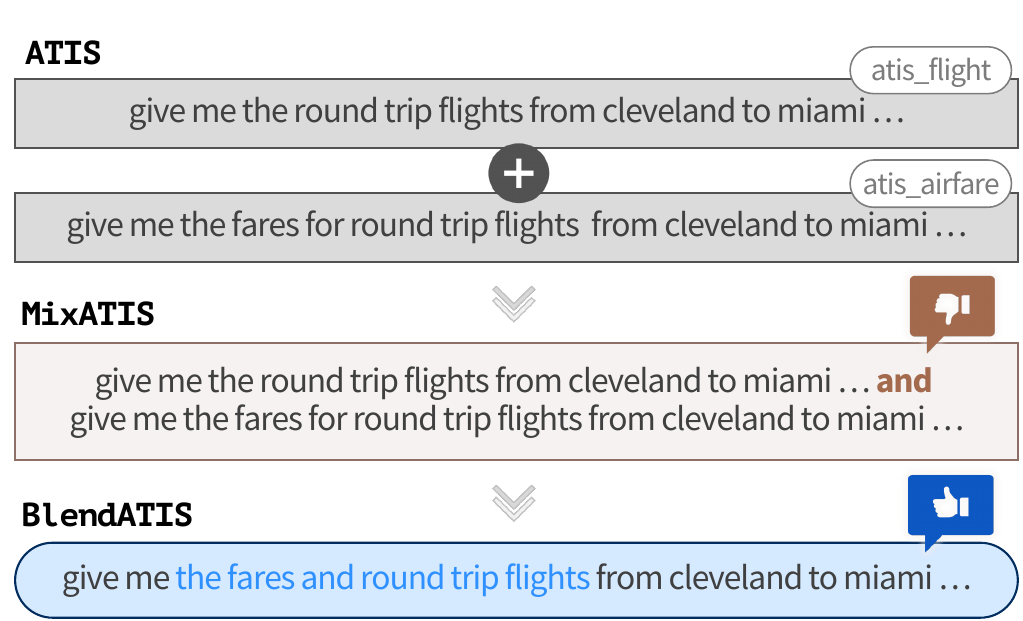} 
        \caption{
        An example that underscores the distinct features of MixX and BlendX.
        In contrast to MixX, which relies on simple concatenations, BlendX steps beyond by simulating more realistic and complex cases often found in real-world conversations.
        }
        \label{fig:comparison_concat}
    \end{center}
\end{figure}

\begin{figure*}[t]
    \begin{center}
        \includegraphics[scale=0.53]{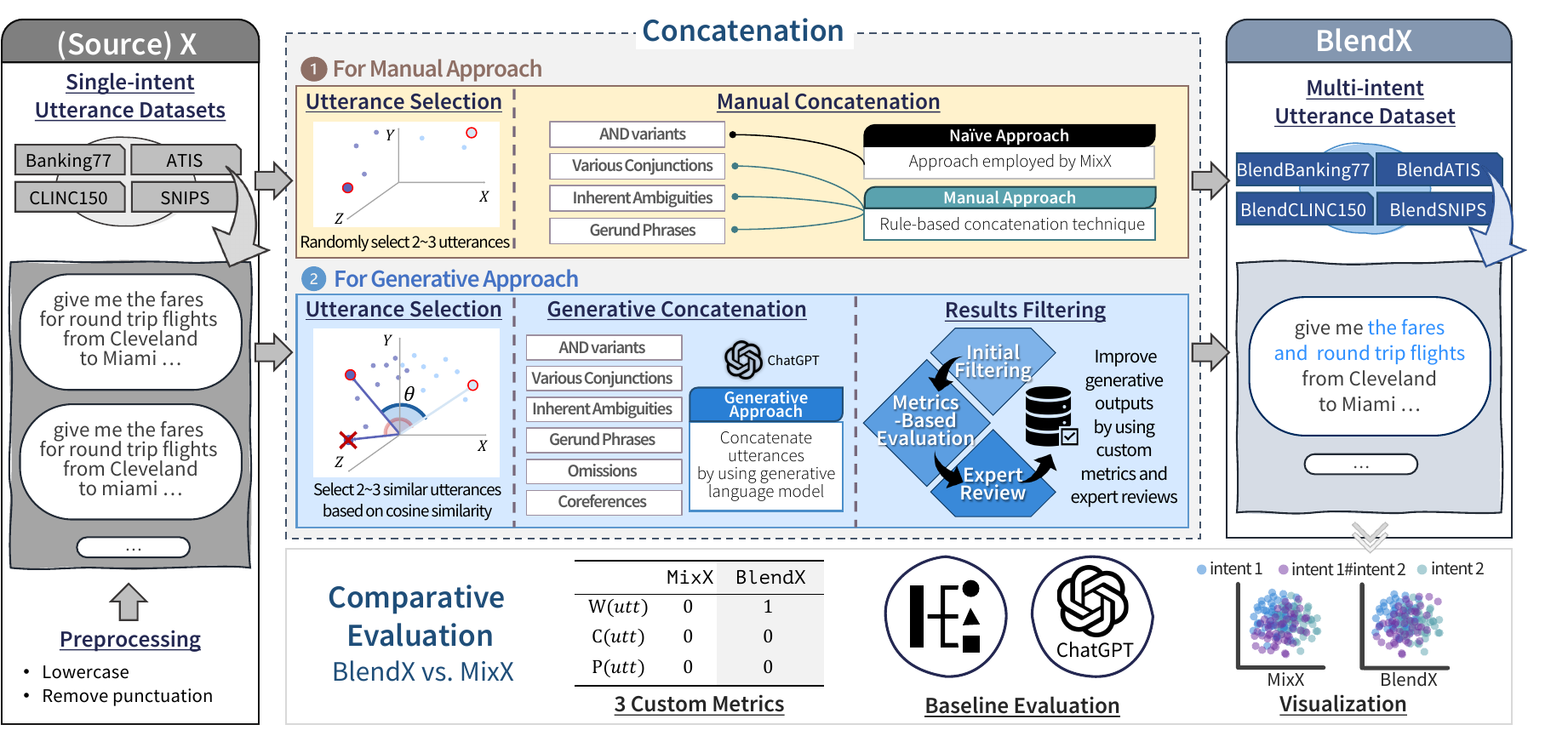} 
        \caption{
        An overview of the BlendX construction framework. 
        Initially, we preprocess four source datasets: ATIS, Banking77, CLINC150, and SNIPS.
        We then select single-intent utterances from these datasets.
        These utterances are combined using both Manual and Generative approaches.
        It is important to note that utterances are kept separate and not mixed across datasets.
        Following the merging process, all resultant datasets are compiled to form BlendX.
        We particularly highlight non-trivial combinations, such as omissions, which are indicated within the blue rounded box on the rightmost side of the framework.
        Finally, BlendX is evaluated using three methods: custom metrics, baseline evaluation, and visualization.
        }
        \label{fig:overall}
    \end{center}
\end{figure*}

Despite the ongoing interest in MID, we find it surprisingly notable that resources supporting this research direction are quite limited.
Most studies on MID rely on two representative datasets, i.e., \textit{MixATIS} and \textit{MixSNIPS} \citeplanguageresource{qin-etal-2020-agif}.
They serve as extensions of the classic single-intent detection datasets---\textit{ATIS} \citeplanguageresource{mansour2021atis} and \textit{SNIPS} \citeplanguageresource{coucke2018snips}---modified to include scenarios that involve multiple intents.

Unfortunately, in spite of its pervasive adoption within the domain, \textbf{MixX}\footnote{A term encompassing both MixATIS and MixSNIPS, as well as the framework used to create the datasets.} \citeplanguageresource{qin-etal-2020-agif} has faced criticism for the simplicity inherent in their construction.
\citet{larson2022survey} highlighted the insufficient diversity in the connectives used to merge multiple utterances into a unified expression in the construction of MixX.
That is, MixX features merely four types of coordinating conjunctions: \texttt{`and'}, \texttt{`and then'}, \texttt{`and also'}, and \texttt{`,(comma)'}, patterns that are susceptible to detection by cutting-edge models.\footnote{For instance, a smart model may exploit the na\"ive patterns to identify the number of intents in an utterance without grasping the utterance's overall semantics.}
Consequently, this casts doubt on the validity of evaluations related to recent MID approaches, especially since the principal components of these evaluations generally lean on the aforementioned datasets, MixX.

In this context, we argue that obvious and urgent needs exist for establishing a more rigorous testbed for MID, as shown in Figure \ref{fig:comparison_concat}.
Remarkably, this comes despite the minimal effort noted in the literature to address the issue.
While recent work on MID largely focuses on devising new methodological schemes---evaluated within fixed, simple environments---we aim to offer orthogonal enhancement to the field by introducing a suite of upgraded datasets, dubbed \textbf{BlendX}.

We explore the limitations present in the current form of MID datasets and propose curated datasets featuring more complex and varied patterns.
Initially, we propose pragmatic rules for manually merging single-intent utterances, utilizing an expanded array of connectors that allow us to diverge from the simple heuristics employed in MixX.

Moreover, we consider the automated concatenation of utterances, facilitated by leveraging OpenAI's ChatGPT\footnote{In particular, we employ \texttt{gpt-3.5-turbo-0613}.}.
We find that, although ChatGPT is versatile, its na\"ive utilization struggles to merge  given utterances while preserving their original intents.
To maximize its efficacy, we introduce a similarity-based strategy for utterance selection, aiding the model to operate in more realistic settings without being excessively challenging.

In addition, we propose three intuitive metrics designed to assess the quality of the constructed datasets.
Our analysis with these metrics demonstrates that BlendX significantly outperforms its predecessors in terms of complexity and diversity.

Lastly, we revisit state-of-the-art MID models, i.e., TFMN \cite{chen-etal-2022-transformer} and SLIM \cite{9747477}, as well as ChatGPT to evaluate their performance on BlendX.
We discover that MID models struggle to adapt to the distinctive patterns present in BlendX, prompting a re-evaluation of the current state in MID literature.
We also provide extensive analysis of BlendX's attributes, shedding light on its unique contributions.
\section{Related Work}
\label{sec:related_work}

\paragraph{Single-Intent Detection Datasets} 
We present datasets for single-intent detection, which serve as the foundation for more complex settings.
One of the classic resources in the field of intent detection is the ATIS dataset \citeplanguageresource{mansour2021atis}, which includes utterances about 26 airline-related intents.\footnote{Indeed, 8 of the original 26 intents in ATIS are multi-intent classes, which we exclude from our analysis.}
Meanwhile, the SNIPS dataset \citeplanguageresource{coucke2018snips} consists of utterances with 7 intents. 
These two datasets have extensions in MID settings, i.e., MixATIS and MixSNIPS \citeplanguageresource{qin-etal-2020-agif}.

Besides ATIS and SNIPS, there exist many other datasets for (single-)intent detection. 
We aim to expand upon these datasets by introducing them to the MID setting, similar to the cases of MixATIS and MixSNIPS.
One of the candidates within the scope of our study is Banking77 \citeplanguageresource{casanueva-etal-2020-efficient}, a dataset that comprises 13,083 customer service queries with 77 labeled intents specific to the banking domain.
Another target of our study is CLINC150 \citeplanguageresource{larson-etal-2019-evaluation}. 
This dataset encompasses 23,700 examples distributed across 150 intents within 10 domains, and offers additional distinct out-of-domain (OOD) instances.\footnote{For our purpose, we exclude data instances with the out-of-domain (OOD) intent, having a total of 150 intents.}

In summary, our work centers on four single-intent datasets---ATIS, SNIPS, Banking77, and CLINC 150---along with their extension into MID environments.\footnote{We leave exploration on other datasets, e.g., HWU64 \citeplanguageresource{liu2019benchmarking}, SLURP \citeplanguageresource{bastianelli-etal-2020-slurp}, and RedWood \citeplanguageresource{larson2022redwood}, as future work.}
In the following, we illustrate the current status and limitations of existing MID datasets. 

\begin{figure*}[t]
    \begin{center}
        \includegraphics[scale=0.7]{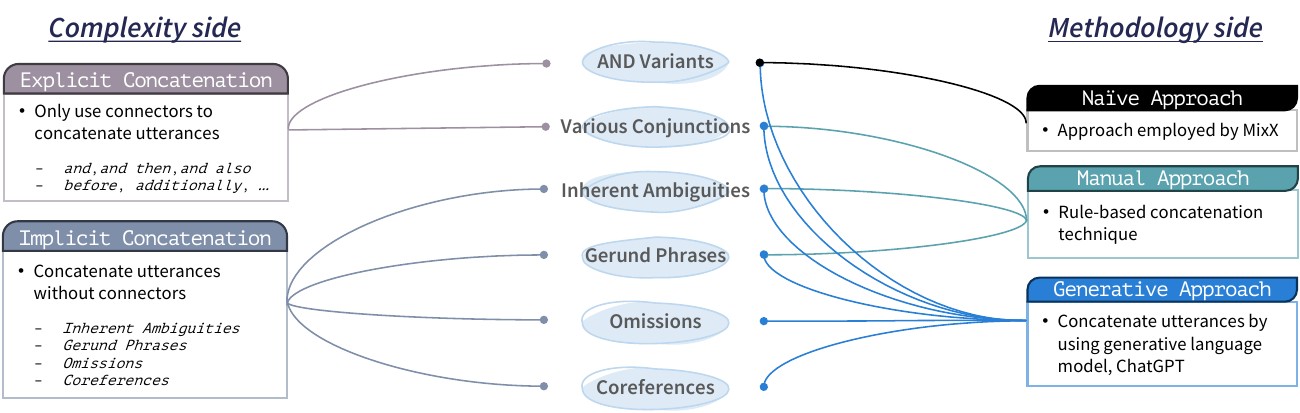} 
        \caption{Illustration of the complexity (Left) and methodology (Right) aspects of concatenation. 
        Each approach triggers a distinct part of the possible variations (Middle) arising in the process of concatenation.}
        \label{fig:concat}
    \end{center}
\end{figure*}

\paragraph{Multi-Intent Detection Datasets} 
\citet{larson2022survey} indicate the scarcity of resources tailored for multi-intent detection.
Notably, MixATIS and MixSNIPS \citeplanguageresource{qin-etal-2020-agif} have played a pivotal role in supporting nearly every experiment in MID.
Since MixATIS and MixSNIPS were both proposed by the same group of researchers \citeplanguageresource{qin-etal-2020-agif}, they share common characteristics.
The datasets consist of utterances with up to three intents, and the distribution of inputs with different intents (1:2:3) is maintained at a ratio of 3:5:2.
Furthermore, they consistently utilize the term \texttt{`and'} (along with its variations) to merge multiple utterances into a unified one.
We also note that the \texttt{`,(comma)'} is used in the datasets exclusively when concatenating three utterances in a row.

These explicit patterns can provide strong cues for models. 
For example, a model might learn to identify the number of intents by either (1) counting the occurrences of the conjunction \texttt{`and'} or (2) recognizing the presence of a \texttt{`,(comma)'} indicating three intents.
If this hypothesis holds true, models trained on MixX may encounter notable performance drops when evaluated on datasets lacking or having fewer clues.
We thus intend to verify our conjecture by introducing a novel suite of datasets equipped with more diverse patterns.

Lastly, the recently introduced dataset named DialogUSR \citeplanguageresource{meng-etal-2022-dialogusr} stands out as a significant resource for MID research.
This dataset is characterized by its provision of pairs consisting of a multi-intent utterance and its corresponding single-intent sub-queries, all annotated by humans.
As a result, it enables models to learn the process of dissecting a multi-intent utterance and accurately extracting the resulting single-intent sub-queries.
However, its dependence on human annotations presents a clear drawback due to the associated costs. 
Furthermore, the dataset is built on the assumption that multi-intent utterances can be completely segmented, which may not hold true in real-world scenarios.
\section{Dataset Construction}
\label{sec:dataset_construction}

We introduce a framework to construct a novel suite of datasets tailored for multi-intent detection, termed \textbf{BlendX}, as illustrated in Figure \ref{fig:overall}. 
Inspired by MixX \citeplanguageresource{qin-etal-2020-agif}, our approach merges utterances from single-intent datasets. 
We broaden the research scope by incorporating datasets such as Banking77 and CLINC150 and by utilizing diverse conjunctions. 
Furthermore, we explore methods to connect utterances, including rule-based ones plus generative models, i.e., ChatGPT.
We also propose simple but effective metrics to validate the quality of the generated datasets.

\begin{table*}
    \scriptsize
    \centering
    \extrarowheight=\aboverulesep
    \addtolength{\extrarowheight}{\belowrulesep}
    \aboverulesep=0pt
    \belowrulesep=0pt
    \setlength{\tabcolsep}{0.45em}
        \begin{tabular}{>{\centering\arraybackslash\columncolor{Gray}}m{1.0cm} x{3.5cm} y{3.5cm} x{4.7cm} y{2.3cm}}
        \toprule
        \textbf{Success} & \textbf{Utterance 1} & \textbf{Utterance 2} & \textbf{Concatenation} & \textbf{Implication} \\
        \midrule
         X & weather for frankfort (\texttt{GetWeather}) & 3 out of 6 for the last album (\texttt{RateBook}) & frankfort's weather gets a 3 out of 6 rating for the last album & Intent (\texttt{RateBook}) removed \\
         & show me the fares for a first class ticket from baltimore to dallas (\texttt{atis\_airfare}) & i would like the time of all flights from san francisco to pittsburgh on sunday (\texttt{atis\_flight\_time}) & show me the fares for a first class ticket from baltimore to dallas and i would like to know all flights from san francisco to pittsburgh on sunday & Intent changed (from \texttt{atis\_flight\_time} to \texttt{atis\_flight}) \\
        & what is mci (\texttt{atis\_abbreviation}) & list la (\texttt{atis\_city}) & combine the sentences: "what is mci?" and "list la". & Failed to merge \\
        \midrule
        O & play the playlist funtime activity 
        (\texttt{AddToPlaylist}) & add adele onto funtime activity playlist
        (\texttt{PlayMusic}) & play the playlist, adding adele to the funtime activity playlist & Gerund phrase \\
        & how do i freeze my bank account	
        (\texttt{account\_blocked}) & do you know why my bank account is frozen (\texttt{freeze\_account}) & how do i freeze my bank account, and do you know why it is frozen & Coreference \\
        \bottomrule
        \end{tabular}
    \caption{Failed and successful results of the Generative Approach and their implications.}
    \label{table:generative}
\end{table*}

\subsection{Concatenation}
\label{subsec:concatenation}
MID datasets are typically created by merging two utterances using connectives. 
However, this fails to encompass the full range of ways people express multiple intents, as they often employ varied connectives or omit them entirely.
To assemble multi-intent utterances with nuanced patterns and to improve upon the rule-based approach suggested by MixX, we view concatenation from two distinct aspects, as in Figure \ref{fig:concat}: 
\textbf{the complexity of concatenation} (explicit or implicit) and \textbf{the methodology of performing concatenation} (whether conducted manually or through tools such as ChatGPT.)

\paragraph{The complexity aspect}
We present two merging methods based on their level of complexity.

\begin{enumerate}
    \item \textbf{Explicit Concatenation:}    
    Conjunctions are explicitly used to concatenate two or more utterances.\footnote{Specifically: \texttt{`and'}, \texttt{`and then'}, \texttt{`and also'}, \texttt{`,(comma)'}, \texttt{`;(semi-colon)'}, \texttt{`or'}, \texttt{`before'}, \texttt{`after'}, \texttt{`additionally'}, \texttt{`finally'}.\label{fn:conjunctions}}
    We refer to the use of the four connectives as outlined in \citetlanguageresource{qin-etal-2020-agif}---\texttt{`and'}, \texttt{`and then'}, \texttt{`and also'}, \texttt{`,(comma)'}---as the \textbf{AND variants} setting.    
    If other conjunctions are employed, we denote it as \textbf{various conjunctions} (see Figure \ref{fig:concat}).
    \item \textbf{Implicit Concatenation:} This approach pursues a seamless blend of utterances, minimizing the apparent usage of conjunctions.
    \citet{meng-etal-2022-dialogusr_} discovered that 62.5\% of the follow-up queries in the dialogues they gathered were either incomplete or incorrectly formulated.
    This inspires us to consider the four following implicit merging patterns.
    \textbf{Inherent ambiguity (conjunction removal)} refers to cases where conjunctions are simply removed from their original positions.
    Despite its simplicity, it effectively reflects the intuition that speakers tend to favor shorter utterances.
    In line with the same philosophy, we also consider \textbf{omissions} and \textbf{coreferences}, where redundant expressions are either eliminated or substituted with pronouns.
    Lastly, we employ \textbf{gerund phrases} (the \texttt{-ing} form of verbs) which are useful for emphasizing concurrency.
    For a clearer understanding of readers, we provide examples of each case in Table \ref{table:concat_and_metrics}.
\end{enumerate}

\paragraph{The methodology aspect}
The remaining issue is how we implement the phenomena we have specified.
While we permit minor adjustments to source sentences during merging, like omissions and coreferences, we aim to retain the sentences' original structure to the greatest extent possible.
In alignment with our mission, we propose two implementation strategies: manual rule-based heuristics and the utilization of ChatGPT.

\begin{enumerate}
    \item \textbf{Na\"ive Approach:}
    It follows the original practices proposed in \citetlanguageresource{qin-etal-2020-agif}.
    It is Explicit Concatenation with the AND variants setting. 
    \item \textbf{Manual Approach:} It expands the Na\"ive Approach with further rule-based techniques. 
    To be specific, it utilizes a broader range of conjunctions (see Footnote \ref{fn:conjunctions}), occasionally skips connectives, and, when feasible, transforms one of the utterances into a gerund phrase.
    \item \textbf{Generative Approach}: Facing the non-trivial challenge of implicit concatenations, we attempt to circumvent this issue with the aid of large language models, especially ChatGPT.
    We instruct the tool to merge utterances while preserving their original intents and structures as much as possible, and we specifically encourage it to avoid using conjunctions such as \texttt{'and'}.
    Figure \ref{figure:prompt_chatgpt_concat} shows the prompt used in the process. 
    We provide three few-shot samples each of both successful and unsuccessful merges.
    More detailed information about the prompt is available in Appendix \ref{subsec:prompt_concat}.
    \label{table:metric_samples}
\end{enumerate}

\begin{figure}[t]
    \setlength{\fboxsep}{7pt} 
    \noindent\fbox{%
        \begin{minipage}{\dimexpr\columnwidth-2\fboxsep-2\fboxrule\relax}
        
            \fontsize{9pt}{11pt}\selectfont 
            \setlength{\parskip}{5pt} 
            \textbf{Instructions:} \textsc{You are a native English speaker. Combine the following sentences as one single sentence naturally.} \\
            \vspace{-5pt}

            \textcolor{commentcolor}{\# $N$ iterations} \\
            \vspace{-5pt}
            \textbf{Example $N$: } [($\mathrm{utt_1}$, $\mathrm{intent_1}$), ..., ($\mathrm{utt_k}$, $\mathrm{intent_k}$)]
            \begin{itemize}
                \item \textbf{Good Answer: } \textsc{[sample answer]} \\\vspace{-3pt}\textcolor{commentcolor}{\# An ideal result of concatenating example utterances}
                \item \textbf{Bad Answer: } \textsc{[sample answer]} \\\vspace{-3pt}\textcolor{commentcolor}{\# An unwanted or incorrect result of concatenating example utterances}
            \end{itemize}

            \textbf{Query: } \textsc{[query]} \\\vspace{-5pt}
            
            \textbf{Answer: } \textsc{[answer]}
        \end{minipage}%
    }
    \caption{
    Prompt design for the Generative Approach. 
    The demonstration section showcases $N$ ($=$ 3)  examples of combining $k$ ($=$ 2 or 3) utterances, featuring both a successful and a failed case. 
    The \textsc{[query]} lists the sentences to be merged.
    ChatGPT performs the merging process by filling in the \textsc{[answer]} part. 
    Blue comments are for illustrative purposes only and are not part of the actual prompt.
    }
    \label{figure:prompt_chatgpt_concat}
\end{figure}

In our preliminary experiments, we discovered that ChatGPT often fails to combine two given sentences with the expected level of creativity and naturalness.
Consequently, contrary to our expectations, the initial results from ChatGPT closely reproduced those from the Manual Approach, which heavily relies on directly using conjunctions.
For reference, Table \ref{table:generative} displays a few examples of failed and successful outputs from ChatGPT.

Nevertheless, we noted some instances where ChatGPT excels in producing high-quality utterances (e.g., the bottom part of Table \ref{table:generative}) that could not be simply achieved through rule-based heuristics.
We highlight that manually designing such examples requires an exceptional level of expertise and effort, and they are not readily achievable even by human annotators.
Therefore, in the subsequent section, we investigate techniques in terms of utterance selection to better condition ChatGPT for creating more reliable data samples.

\begin{table*}
    \scriptsize
    \centering
    \extrarowheight=\aboverulesep
    \addtolength{\extrarowheight}{\belowrulesep}
    \aboverulesep=0pt
    \belowrulesep=0pt
        \begin{tabular}{l a a c c a a c c}
        \toprule
            \multirow{2}{*}{\shortstack[c]{\textbf{Metric}}} & \multicolumn{2}{a}{\textbf{SNIPS}} & \multicolumn{2}{c}{\textbf{ATIS}} & \multicolumn{2}{a}{\textbf{Banking77}} & \multicolumn{2}{c}{\textbf{CLINC150}} \\
            & \textbf{Random} & \textbf{Sim.} & \textbf{Random} & \textbf{Sim.} & \textbf{Random} & \textbf{Sim.} & \textbf{Random} & \textbf{Sim.} \\
        \midrule
            Cosine sim. & 0.105 & 0.746 & 0.214 & 0.758 & 0.212 & 0.748 & 0.093 & 0.749   \\
            Error rate $(\downarrow)$ & 16\% & \textbf{14\%} & 41\% & \textbf{10\%} & 22\% & \textbf{9\%} & 19\% & \textbf{13\%} \\
        \midrule
            $\mathrm{W}(utt, 2) (\uparrow)$   & 27.38\% & \textbf{44.87}\% & 10.17\% & \textbf{27.78}\% & \textbf{34.62\%} & 30.77\% & 30.86\% & \textbf{31.03\%}  \\
            $\mathrm{C}(utt, 2) (\uparrow)$ & \textbf{8.33\%} & 1.28\% & 3.39\% & \textbf{4.44\%} & \textbf{28.21\%} & 15.38\% & \textbf{25.93\%} & 3.45\% \\
            $\mathrm{P}(utt, 2) (\uparrow)$ & 3.57\% & \textbf{10.26\%} & 1.69\% & \textbf{12.22\%} & 10.26\% & \textbf{20.88\%} & 3.70\% & \textbf{14.94\%} \\
        \bottomrule
        \end{tabular}
    \caption{
    Comparison of Random and Similarity-Based (Sim.) utterance selection across datasets when applied to ChatGPT.
    We find that Sim. leads to a reduced error rate in ChatGPT's data generation.}
    \centering
    \label{table:selection_metric_table}
\end{table*}

\subsection{Utterance Selection}
\label{subsec:utterance_selection}
In the original MixX setting, source utterances for concatenation are randomly chosen without specific criteria. 
Our study introduces an additional selection approach rooted in utterance embedding similarity. 
For a given pair of utterances, we employ SBERT \cite{reimers-gurevych-2019-sentence} to compute the cosine similarity of their sentence embeddings (i.e., the \texttt{[CLS]} token embeddings). 
We only include the sentence pair in data construction only if their score exceeds a certain threshold $\tau$.\footnote{$\tau$ is empirically set to 0.7. 
For some cases, we set it to 0.4 to achieve the proper training-dev-test split ratio. 
Table \ref{table:selection_metric_table} ensures that we use semantically similar utterances rather than random samples even when $\tau$=0.4.}
When combining three utterances into one, we choose the set where all possible pairs surpass the threshold.

Consequently, we propose using the two following methods for utterance selection. 
\begin{enumerate}
    \item \textbf{Random Selection:}
    As in MixX, utterances are randomly chosen without any specific rule.
    Random Selection accounts for the potential scenarios in spoken language, where inputs are often noisy or ungrammatical.
    \item \textbf{Similarity-Based Selection:}
    Utterance pairs are selected based on their similarity scores.
\end{enumerate}
We utilize Random Selection for the Na\"ive and Manual Approach, while the Gerative Approach is combined with the Similarity-Based Selection.\footnote{
Prior experiments suggest the Manual Approach is not significantly impacted by utterance selection.}
To justify our approach, Table \ref{table:selection_metric_table} presents an experiment for demonstrating the effectiveness of Similarity-Based Selection when integrated with ChatGPT.
\begin{table*}
    \scriptsize
    \centering
    \extrarowheight=\aboverulesep
    \addtolength{\extrarowheight}{\belowrulesep}
    \aboverulesep=0pt
    \belowrulesep=0pt
    \begin{tabular}{l b c c c}
        \toprule
            \textbf{Utterance 1} & \multicolumn{4}{b}{play my 88 keys playlist (\texttt{PlayMusic})} \\
            \textbf{Utterance 2} & \multicolumn{4}{b}{add another song to my 88 keys playlist (\texttt{AddToPlaylist})} \\ 
        \bottomrule
        \toprule
            \textbf{Strategies }& \textbf{Concatenation Results} & $\mathrm{W}(utt, 2)$ & $\mathrm{C}(utt, 2)$ & $\mathrm{P}(utt, 2)$ \\
        \midrule
            \textbf{Explicit Concatenation} & play my 88 keys playlist \textbf{and also} add another song to my 88 keys playlist & 0 & 0 & 0 \\
        \midrule
            \textbf{Implicit Concatenation} & & & & \\ 
            Inherent Ambiguities & play my 88 keys playlist add another song to my 88 keys playlist & 1 & 1 & 0 \\ 
            Gerund Phrases & add another song to my 88 keys playlist playing it  & 1 & 1 & 1 \\ 
            Omissions & play my 88 keys playlist and add another song & 1 & 0 & 0 \\ 
            Coreferences & play my 88 keys playlist and add another song to it & 1 & 0 & 1 \\ 
        \bottomrule
    \end{tabular}
    \caption{Various concatenation classes, accompanied by their examples and respective metric values.}
    \centering
    \label{table:concat_and_metrics}
\end{table*}

\subsection{Three Custom Metrics} \label{subsec:metric}
Beyond formulating the data construction process for MID, our goal also includes developing a method to quantitatively assess the quality of the generated datasets.
To achieve this, we introduce three custom metrics, based on the hypothesis that the complexity and diversity of merged instances can be measured by analyzing variances in word count, conjunction usage, and pronoun frequency before and after concatenation.

For instance, consider a scenario in which the utterances in Table \ref{table:concat_and_metrics}, \texttt{`play my 88 keys playlist'} and \texttt{`add another song to my 88 keys playlist'} are merged to form \texttt{`add another song to my 88 keys playlist playing it'}(Gerund Phrase). 
The number of words of the two original utterances are 5 and 8, respectively.
After concatenation, the number becomes 10.
We can apply the same calculation for conjunctions and pronouns, realizing the fact that the concatenated utterance has three fewer words, no conjunctions, one more pronoun than the original utterances.
This example provides insight into numerically estimating the degree of linguistic transformations resulting from concatenation.

Suppose a merged utterance $utt$, which is the concatenation of $n$ utterances out of a total of $m$ candidate utterances, denoted as $utt_1, utt_2, \cdots, utt_m$, where each individual utterance $utt_i$ is labeled with an intent $intent_i$. 
Let $|utt|_{x}$ be a function which counts the number of $x$ in $utt$. 
The element $x$ can represent any linguistic feature within the utterance, such as words, conjunctions, or pronouns, depending on the context.
Here, $n$ represents the number of utterances that were actually concatenated to form $utt$, and $m$ represents the total number of candidate utterances available for concatenation.

Let $\mathrm{W}(utt, n)$ be a function to count and subtract the number of words in $utt$ after and before concatenation, and indicate if the value is less than 1 or not.
The exact definition of $\mathrm{W}(utt, n)$ is as follows: 
\begin{equation}
\mathrm{W}(utt, n)\\
\overset{\underset{\mathrm{def}}{}}{=}\mathbf{1}_{\mathbb{Z}-\mathbb{N}}\Bigl(|utt|_{word}-\sum_{i=1}^n |utt_i|_{word}\Bigr).    
\end{equation}

$\mathrm{C}(utt, n)$ is similar to $\mathrm{W}(utt, n)$, but it counts conjunctions instead:
\begin{equation}
\mathrm{C}(utt, n)\\
\overset{\underset{\mathrm{def}}{}}{=}\mathbf{1}_{\mathbb{Z}-\mathbb{N}}\Bigl(|utt|_{conj}-\sum_{i=1}^n |utt_i|_{conj}\Bigr).
\end{equation}

$\mathrm{P}(utt, n)$ counts and subtracts the number of pronouns in $utt$ after and before concatenation, and indicates if the value is more than 0 or not:
\begin{equation}
\mathrm{P}(utt, n)\\
\overset{\underset{\mathrm{def}}{}}{=}\mathbf{1}_{\mathbb{N}}\Bigl(|utt|_{pron}-\sum_{i=1}^n |utt_i|_{pron}\Bigr).
\end{equation}

These metrics help us identify patterns in the integrated utterances, offering insights into the complexity of the concatenation procedure. 
If each metric becomes 1, it indicates an omission of some words, the inclusion of conjunctions, and the presence of pronouns.
As such, we expect that in an ideal scenario, all metrics would converge to one.

While our metrics can serve as indicators of the complexity inherent in a concatenation process, they are not perfect; they cannot ensure a fully accurate and complete concatenation that adheres to both semantic and syntactic norms.
Still, we claim that the proposed metrics can effectively serve as a proxy for measuring the quality of the merging process, as evidenced in the following sections.

\begin{table*}
    \scriptsize
    \centering
    \extrarowheight=\aboverulesep
    \addtolength{\extrarowheight}{\belowrulesep}
    \aboverulesep=0pt
    \belowrulesep=0pt
    \setlength{\tabcolsep}{0.5em}
    \begin{tabular}{l a a a c c c a a a c c c}
    \toprule
        \multirow{2}{*}{\shortstack[c]{\textbf{Metric}}} & \multicolumn{3}{a}{\textbf{SNIPS}} & \multicolumn{3}{c}{\textbf{ATIS}} & \multicolumn{3}{a}{\textbf{Banking77}} & \multicolumn{3}{c}{\textbf{CLINC150}} \\
        & \textbf{Na\"ive} & \textbf{Manual} & \textbf{Generative} & \textbf{Na\"ive} & \textbf{Manual} & \textbf{Generative} & \textbf{Na\"ive} & \textbf{Manual} & \textbf{Generative} & \textbf{Na\"ive} & \textbf{Manual} & \textbf{Generative} \\
    \midrule
        $\mathrm{W}(utt, 2) (\uparrow)$  & 0\% & \textbf{37\%} & 29\% & 0\% & \textbf{36\%} & 18\% & 0\% & \textbf{46\%} & 37\% & 0\% & \textbf{48\%} & 28\% \\
        $\mathrm{C}(utt, 2) (\uparrow)$ & 0\% & \textbf{56\%} & 10\% & 0\% & \textbf{52\%} & 15\% & 0\% & \textbf{50\%} & 27\% & 0\% & \textbf{56\%} & 32\% \\
        $\mathrm{P}(utt, 2) (\uparrow)$ & 0\% & 0\% & \textbf{7\%} & 0\% & 0\% & \textbf{8\%} & 0\% & 0\% & \textbf{13\%} & 0\% & 0\% & \textbf{6\%} \\
    \bottomrule
    \end{tabular}
\caption{
Comparative analysis of the three concatenation approaches: Na\"ive, Manual, and Generative. 
Notably, the Manual method demonstrates pronounced efficiency in reducing utterance length.}
\centering
\label{table:concatenation_metric_table}
\end{table*}

\paragraph{Analysis of utterance selection with ChatGPT} 
Let us first revisit the prior experiment associated with Table \ref{table:selection_metric_table}.
We select 100 pairs of utterances using each utterance selection strategy, resulting in a notable disparity in the average cosine similarities, as shown in the first row of Table \ref{table:selection_metric_table}.
The \textbf{Error Rate} correlates with the concept of \textit{intent distortion}, which refers to the occurrence of either removal or alteration of the original intentions during the merging process. 
This rate is calculated by taking the number of merged utterances that have \textit{intent distortion} and dividing it by the total number of utterances that were reviewed.
The row for Error Rate in the table indicates that, when combined with Similarity-Based Selection, ChatGPT makes significantly fewer errors (ranging from 2\% to 31\%) in merging two given sentences.

Again in Table \ref{table:selection_metric_table}, we provide analysis with our novel metrics. 
When comparing the results of Similarity-Based Selection with those of Random Selection, we note an increase in the usage frequency of pronouns across datasets. Conversely, there is a decline in utterance word count for all datasets, except Banking77.\footnote{
Banking77's longer, multi-sentence utterances led ChatGPT to often use simple \texttt{`and'} for concatenation.}
These findings imply that Similarity-Based Selection results in more omissions and coreferences which are desirable.

Notably, utterances that feature omissions or coreferences tend to be more ambiguous. 
As such, the use of conjunctions, including \texttt{`and'}, becomes essential to maintain semantic clarity. 
Consequently, it is expected that the percentage of $\mathrm{C}(utt,2)$ will decrease more with Similarity-Based Selection than with Random Selection.
Note that higher values for all metrics are indicative of a higher likelihood of achieving a desirable merging process.
Indeed, for connections of greater complexity, it is anticipated that all metrics will exhibit increased values. 
The sole exception to this rule is in instances of coreference or omission, where the value may decrease to $\mathrm{C}(utt,2)$.

\paragraph{Analysis of concatenation approaches} 
Finally, we compare the effectiveness of three concatenation techniques, i.e., {Na\"ive, Manual, and Generative, with the new metrics.
For evaluation, we produce 100 instances using each approach.
Results are listed in Table \ref{table:concat_and_metrics}.
We discover that only the Manual and Generative methods enable implicit concatenation. Additionally, we find that Naïve concatenation, unsurprisingly, does not result in shorter concatenated utterances; it neither introduces pronouns nor omits conjunctions. 
As shown in Table \ref{table:concatenation_metric_table}, Manual demonstrates its effectiveness, consistently reducing the length of concatenated utterances by 1.2 to 2 times compared to ChatGPT. 
Furthermore, Manual tends to use fewer conjunctions. 
Surprisingly, we observe that ChatGPT favors more conjunctions than expected. 
Upon examining metric $\mathrm{C}(utt, 2)$ in the table,  it is evident that ChatGPT employs the conjunction \texttt{`and'} in 40\% to 72\% of its concatenations, indicating a possible bias towards simpler concatenation strategies. 
Interestingly, even when explicitly instructed to avoid using \texttt{`and'}, as detailed in Appendix \ref{subsec:prompt_concat}, ChatGPT often disregards this directive.
\subsection{Dataset Details}  \label{subsec:dataset_details}

\begin{table}[t]
    \footnotesize
    \centering
    \setlength{\tabcolsep}{0.3em}
    \begin{tabular}{l c c c c c}
        \toprule
            \textbf{Dataset} & \textbf{Intents \#} & \textbf{Training} & \textbf{Dev} &\textbf{Test} & \textbf{Total} \\
        \midrule
            SNIPS & 7 & 50,625 & 2,613 & 2,615 & 55,853 \\
            ATIS & 18 & 20,250 & 1,125 & 1,125 & 22,500 \\
            Banking77 & 77 & 36,390 & 2,009 & 2,021 & 40,420 \\
            CLINC150 & 147 & 54,896 & 2,889 & 2,977 & 60,762 \\
        \bottomrule
    \end{tabular}
    \caption{Statistics of the constituents of BlendX.}
    \centering
    \label{table:BlendX_statistics}
\end{table}

To summarize this section, we introduce \textbf{BlendX}, a collection of enhanced multi-intent datasets, shaped by our concatenation strategies and further validated using our three newly developed metrics.
We apply our framework to four single-intent datasets: ATIS, SNIPS, Banking77, and CLINC150---resulting in BlendATIS, BlendSNIPS, BlendBanking77, and BlendCLINC150.
The statistics of BlendX are listed in Table \ref{table:BlendX_statistics}.

In the preprocessing step, similar to  MixX \citeplanguageresource{qin-etal-2020-agif}, BlendX ensures that each intent has equal number of utterances in the datasets. 
BlendX may also necessitate duplicating certain utterances, maintaining a ratio of single-, double-, and triple-intent utterances at 3:5:2.
We use single-intent datasets without rectifying their internal errors, and exclude utterances that do not correspond to specific requests.\footnote{e.g., the utterances in CLINC150 whose intents are \texttt{yes}, \texttt{no}, or \texttt{maybe}.} 
For the case of ATIS, each subset of data (training, dev, and test) contains unique types of intents. 
When adapting this to BlendATIS, we retain these characteristics, ensuring it remains relatively more challenging.

We develop the final version of BlendX using both the Manual and Generative concatenation approaches.
For the Manual approach, we keep a balanced ratio for the data instances corresponding to the settings of AND variants, various conjunctions, inherent ambiguities, and gerund phrases, each at 1:1:1:1.
Additionally, using the Generative approach, we create data instances that amount to half of those produced with the AND variants configuration.
They encompass all potential variations of concatenation, with the expectation that they foster natural and intuitive constructions, potentially featuring omissions and coreferences.

\paragraph{Enhancing data quality through data filtering} 
Through the filtering process, we successfully eliminate errors similar to those highlighted in Table \ref{table:generative}. 
This process actively exploits the three metrics proposed in \S\ref{subsec:metric}. 
Initially, we remove clear failures generated by ChatGPT, such as explicit mentions of an intent label or unnecessary punctuation. 

Subsequently, we check all concatenated sentences using the proposed metrics, paying special attention to and filtering out instances where there is a significant discrepancy before ($ \sum_{i=1}^n |utt_i|_x $) and after concatenation ($|utt|_{x}$). 
For example, a substantial increase in word count may indicate unnecessary paraphrasing by ChatGPT, while a significant decrease might suggest overlooked utterances during concatenation. 
Both $\mathrm{C}(utt, n)$ and $\mathrm{P}(utt, n)$ are also subjected to similar filtering logic. 

Lastly, these filtered instances are reviewed by three human experts, with a focus on excluding instances only if they compromise the intended meaning. 
Instances categorized under \textit{intent removed}, \textit{intent changed}, and \textit{failed to merge} were identified and removed due to their significant deviation from the original intent, as detailed in Table \ref{table:generative}. 
This filtering process significantly enhances the quality of our datasets.
\section{Experiments and Analysis}
\label{sec:experiments_analysis}

We conduct an evaluation of several MID methods based on BlendX, as well as the original MixX. 

\paragraph{Methods} 
In recent years, the joint learning of intent-detection and slot filling has emerged as the \textit{de facto} standard for these tasks.
Given the MixX framework's dual provision of multi-intent and slot information for each utterance, this trend has similarly permeated MID research. \cite{Cheng_2023,Tu_2023,chen-etal-2022-transformer,xing-tsang-2022-co,9747843,qin-etal-2021-gl,jiang-etal-2023-spm}
However, given that our emphasis in this study is solely on (multi-)intent detection, we make modest modifications to the two most predominant supervised methods to suit our objective:
\begin{itemize}
    \item \textbf{TFMN} \cite{chen-etal-2022-transformer,Cheng_2023}: 
    This approach first predicts the number of intents, denoted as $k$, in a multi-intent utterance. 
    Subsequently, it yields the top-$k$ options based on the predicted probability distribution.
    We used a variant of the original TFMN method that considers only utterance-level supervision instead of token-level, as annotating utterances from the Generative approach with token-level labels is challenging.\footnote{Since we focus on (multi-)intent detection, we find it reasonable to consider only sentence-level supervision.}
    \item \textbf{SLIM} \cite{9747477}: 
    This method decomposes multi-label classification into a series of binary classifications. 
    Specifically, it gauges the likelihood of each intent using the \texttt{sigmoid} function and then collects the intents whose probability exceeds a given threshold.
\end{itemize}

Furthermore, we also adopt ChatGPT (\texttt{gpt-3.5-turbo-0613}) as an extra baseline.
Figure \ref{figure:prompt_chatgpt_figure} illustrates how we design prompts for facilitating in-context learning for MID.
Note that the utilization of ChatGPT contrasts with the predominantly supervised fine-tuning approaches.
This sheds light on the potential of generative approaches to tackle MID with little to no examples provided.

\begin{figure}[t]
    \setlength{\fboxsep}{7pt} 
    \noindent\fbox{%
        \begin{minipage}{\dimexpr\columnwidth-2\fboxsep-2\fboxrule\relax}
        
            \fontsize{9pt}{11pt}\selectfont 
            \setlength{\parskip}{5pt} 
            \textbf{Instructions:} \textsc{You are an Intent Detection Model on single utterance. Detect single or more intent(s) of each utterance UP TO 3.} \\\vspace{-5pt}
            
            \textbf{Example: } \textsc{[$k$ pairs of a multi-intent utterance and its corresponding intents]} \\\vspace{-5pt}

            \textbf{Query: } \textsc{[query]} \\\vspace{-5pt}
            
            \textbf{Answer: } \textsc{[answer]}           
               
        \end{minipage}%
    }
    \caption{Prompt design for solving MID with in-context learning. 
    We employ the few-shot setting where $k$ multi-intent utterances are provided, each associated with up to three intents. 
    The \textsc{[Answer]} part is filled in to predict the intent of \textsc{[Query]}. 
    Appendix \ref{subsec:prompt_evaluate} presents detailed illustrations.}
    \label{figure:prompt_chatgpt_figure}
\end{figure}

\begin{table*}
    \scriptsize
    \centering
    \extrarowheight=\aboverulesep
    \addtolength{\extrarowheight}{\belowrulesep}
    \aboverulesep=0pt
    \belowrulesep=0pt
    \begin{tabular}{caaccccc}
        \toprule
            \multirow{2}{*}{\shortstack[c]{\textbf{Model}}} & \multicolumn{2}{a}{\textbf{Split}} & \multicolumn{4}{c}{\textbf{Dataset (Metric: Accuracy)}} \\
            & \textbf{Training} & \textbf{Test} & \textbf{SNIPS} & \textbf{ATIS} & \textbf{Banking77} & \textbf{CLINC150}\\
        \midrule
        \multirow{3}{*}{\shortstack[c]{TFMN}} & MixX & MixX & 95.68* $\pm 0.57$ & 77.98* $\pm 0.57$ & 76.61 $\pm 1.17$ & 85.88 $\pm 1.03$ \\
        & MixX & BlendX & 52.51 $\pm 1.86$ & 42.51 $\pm 1.48$ & 37.31 $\pm 0.81$ & 42.45 $\pm 2.40$ \\
        & BlendX & BlendX & 94.93 $\pm 0.85$ & 76.50 $\pm 0.83$ & 63.99 $\pm 0.81$ & 77.96 $\pm 0.82$ \\ 
        \midrule
            \multirow{3}{*}{\shortstack[c]{SLIM}} & MixX & MixX & 95.97* $\pm 0.23$ & 77.10* $\pm 0.28$ &  83.71 $\pm 0.88$ & 88.67 $\pm 0.56$ \\
            & MixX & BlendX & 93.51 $\pm 0.18$ & 72.80 $\pm 1.48$ &  69.89 $\pm 0.46$ & 73.39 $\pm 2.46$ \\
            & BlendX & BlendX & 95.73 $\pm 0.86$ & 76.92 $\pm 0.84$ & 75.30 $\pm 0.71$ & 85.62 $\pm 0.51$ \\ 
        \midrule
            \multirow{2}{*}{\shortstack[c]{gpt-3.5-turbo}} & - & MixX & 81.68 & 40.30 & 30.90 & 49.22 \\
            & - & BlendX & 76.18 & 38.84 & 22.67 & 37.55 \\
        \bottomrule 
    \end{tabular}
    \caption{
    Evaluation of three competitive MID models on MixX and BlendX.
    The reported numbers represent the averages and standard deviations from five distinct executions.
    The symbol $*$ indicates numbers derived from our re-implementation, where we have specifically excluded joint learning with slot filling.
    }
    \label{table:table_main}
    \centering
\end{table*}

\paragraph{Results} 
The main results are listed in Table \ref{table:table_main}. 
First, we reaffirm the widely recognized observation that current supervised methods show reasonable performance when both trained and evaluated with MixX.
Yet, the narrative shifts significantly when these models are evaluated on our new datasets.
In this configuration (training: MixX \& test: BlendX), where the test data distribution deviates from the training distribution, every model shows a significant performance drop, with some declining by up to 40\%.
This outcome suggests that the original MixX may lack the complexity required to comprehensively evaluate the abilities of MID methods.

On the other hand, while transitioning the training data from MixX to BlendX does lead to some performance recovery on the test set, the results do not match the original performance observed when evaluated on MixX. 
This implies that BlendX intrinsically possesses greater complexity, making it more challenging to master.

Additionally, we find that ChatGPT's performance on MID datasets is subpar. 
This indicates that despite ChatGPT's adaptability, more work is needed to optimize its application for this task.

\begin{figure}[t]
\begin{center}
\includegraphics[scale=0.53]{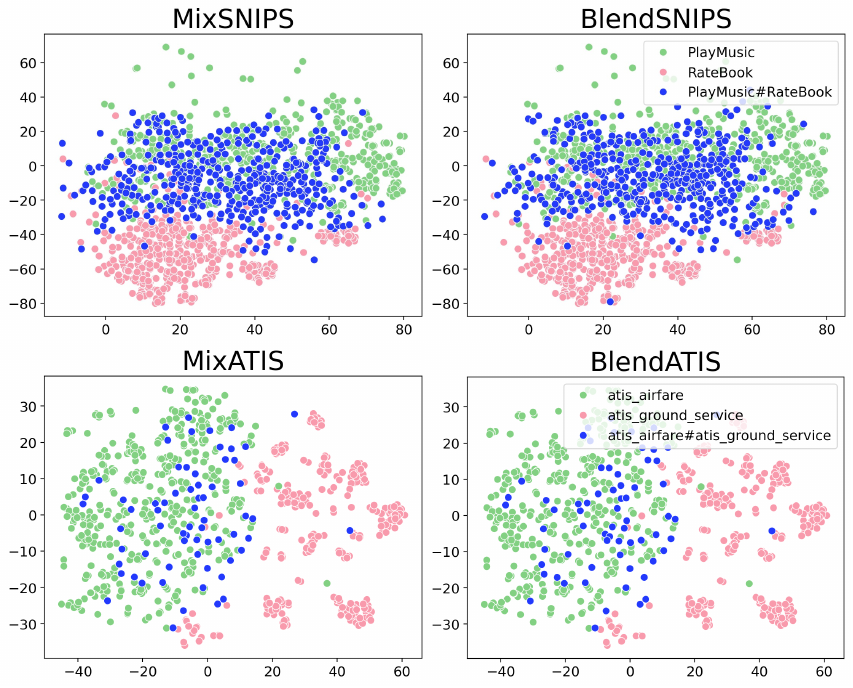} 
\caption{Visualization of MixX (Left) and BlendX (Right) in the $\mathbb{R}^2$ space (datasets: SNIPS, ATIS).}
\label{fig:visualization}
\end{center}
\end{figure}

\paragraph{Visualization}
Figure \ref{fig:visualization} shows data samples from SNIPS and ATIS. 
They are embedded with SBERT and projected into a 2-dimensional space via t-SNE.
In the upper part of the figure, the green and pink dots correspond to utterances whose intents are \texttt{PlayMusic} and \texttt{RateBook}, respectively. 
Meanwhile, the blue dots represent multi-intent utterances with \texttt{PlayMusic\#RateBook}, sourced from MixSNIPS and BlendSNIPS.
The two graphs show nearly identical distributions, implying that although BlendX utterances are expected to be more diverse and noisy, they still occupy a space where data points capture the semantics of both intents.

\begin{table}[t]
\scriptsize
\centering
\extrarowheight=\aboverulesep
\addtolength{\extrarowheight}{\belowrulesep}
\aboverulesep=0pt
\belowrulesep=0pt
    \begin{tabular}{a c c c c}
        \toprule
            \textbf{Data Split} & \multicolumn{4}{c}{\textbf{Testset: BlendX (Metric: Accuracy)}} \\
            \textbf{Generated By} & \textbf{SNIPS} & \textbf{ATIS}& \textbf{Banking77}& \textbf{CLINC150} \\
        \midrule
            Na\"ive & 95.32 & 73.23 & 62.30 & 80.73 \\
            Manual & 25.32 & 42.40 & 8.05 & 25.73 \\
            Generative & 81.58 & 53.93 & 27.95 & 60.17 \\
        \bottomrule
    \end{tabular}
\caption{
Experiments with different subsets of BlendX, grouping data instances by the method used for their creation.
TFMN, trained with MixX, is evaluated on these subsets. 
It is observed that the Manual method contributes the most complexity and difficulty to BlendX.
}
\label{table:ablation_study}
\centering
\end{table}

\paragraph{Ablation study for concatenation methods}
To evaluate the impact of different merging methods within BlendX, we categorize subsets based on the concatenation approach and measured their accuracy using TFMN. 
These results are presented in Table \ref{table:ablation_study}. 
We observe that the subset generated by Manual is consistently more complex than the one created using the Na\"ive method. 
Comparatively, the subset merged via the Generative approach displays greater complexity than the case of Na\"ive but is less intricate than the case of Manual approach. 
These findings indicate that the Manual approach is a highly effective method of construction, facilitating both explicit and implicit concatenation. 
On the other hand, the Generative method tends to yield utterances akin to explicit concatenation, albeit with occasional instances of creativity.
\section{Conclusion}
\label{sec:conclusion}

We highlight the disparity between utterances in class MID datasets and ones existing in more complex, real-world scenarios.
To bridge this gap, we introduce BlendX, a novel suite of multi-intent datasets.
We present three key contributions. 
First, we have transcended the traditional approach by presenting 3 novel concatenation approaches: Na\"ive, Manual, and Generative.
Second, we exhibit the effectiveness of a similarity-based strategy for sentence selection, especially when using this to augment the generative quality of ChatGPT.
Third, we have devised 3 statistical metrics to validate the quality of BlendX.
With extensive experiments, we verify that BlendX provides more challenging environments for MID.
We believe BlendX would facilitate more principled future research in the field.
\section{Limitations}

We believe that the release of our multi-intent dataset framework, known as BlendX, represents a significant advancement in the field of multi-intent detection. However, we have identified several limitations in this work's current state and outline potential future directions to address these limitations.

A prominent issue in our work is our concentration on the MID problem, overlooking the task of slot filling. 
The challenge becomes particularly critical when utterances are modified after concatenation, emphasizing the need for more in-depth exploration in future research to accommodate the slot-filling task. 
Challenges seen in single-intent datasets, such as those in Banking77 with overlapping intents \cite{ying-thomas-2022-label}, continue to exist and highlight the need for more comprehensive dataset refinement strategies.
In Manual approach, we employed a variety of conjunctions; however, there is potential to improve creating more natural utterances by taking into account the relationships between sentences.
The metrics we have developed, focusing on utterance length, conjunctions, and pronouns, do not fully represent linguistic complexity, necessitating further enhancements. 

In conclusion, while our dataset marks a notable advancement in MID, several challenges persist. 
Addressing these will not only amplify our contributions but also serve the broader TOD field.

\section{Acknowledgements}
This work was supported by Hyundai Motor Company and Kia.
This work was supported by Institute of Information \& communications Technology Planning \& Evaluation (IITP) grant funded by the Korea government(MSIT) (No.2020-0-01373, Artificial Intelligence Graduate School Program (Hanyang University)).
This work was supported by Institute of Information \& communications Technology Planning \& Evaluation (IITP) under the artificial intelligence semiconductor support program to nurture the best talents (IITP-(2024)-RS-2023-00253914) grant funded by the Korea government(MSIT).

\section{Bibliographical References}\label{sec:reference}
\bibliographystyle{lrec-coling2024-natbib}
\bibliography{lrec-coling2024}

\section{Language Resource References}\label{lr:ref}

\bibliographystylelanguageresource{lrec-coling2024-natbib}
\bibliographylanguageresource{languageresource}

\appendix
\section*{Appendix}

\section{Prompt Details}
\subsection{For Concatenation}
\label{subsec:prompt_concat}
Table \ref{table:appendix_concatenation_prompt} depicts the detailed prompts of concatenating multiple utterances using ChatGPT. 
When two or three sentences are given as input, the prompt works to combine them into a single sentence while maintaining the essence of each sentence, mirroring the exemplary good answers. 
Despite numerous constraints, the generative approach often struggles to achieve concatenation, as shown in Table \ref{table:generative}.
We experimented with a variety of prompts, including adding or excluding components, in an effort to optimize the concatenation process. 
The term [Bad Answer] indicates an answer is not recommended, not necessarily incorrect.
However, the variations in prompts yielded negligible differences in the quality of the results.

\begin{table*}
    \small
    \centering
    \renewcommand{\arraystretch}{1.5}
    \setlength{\tabcolsep}{0.5em}
    
    \begin{tabular}{>{\raggedright\arraybackslash}p{15cm}}
        \toprule
        You are a native English speaker. \\
        
        \textbf{[Task Definition]} Combine 2 or 3 utterances as one single utterance.\\
        
        \textbf{[Goal]} The focus is on creating a single utterance that captures the essence of both ideas without unnecessary redundancy.\\
        
        \textbf{[Instructions]}
        \begin{itemize}[leftmargin=10mm, nosep]
            \item Avoid adding just punctuation.
            \item Don't paraphrase.
            \item Don't compromise the meaning of each utterance.
            \item Don't replace numbers with radix.
            \item Maintain the intent of each utterance.
            \item Don't forget that if a utterance starts with a verb, it's a statement.
            \item Do NOT use conjunctions like `and'.
            \item Don't print intent directly. 
        \end{itemize} \\
        \arrayrulecolor{lightgray}\midrule\arrayrulecolor{black} 
        
        \textcolor{commentcolor}{\# $N$ iterations}\\
        \textbf{[Example 1]} 
        \begin{quote}
            
        play my 88 keys playlist (\texttt{PlayMusic}) + add another song to my 88 keys playlist (\texttt{AddToPlaylist}) 
        
        {[Good Answer]} while playing my 88 keys playlist, add another song to it. 
        
        {[Bad Answer]} Play my 88 keys playlist and also add another song to my 88 keys playlist. 
        
        \end{quote}
        ... \\
        
        \arrayrulecolor{lightgray}\midrule\arrayrulecolor{black}
        
        \textbf{[Query]} Combine the following utterances naturally.
        
        Inside the parentheses is the intent of each utterance: $\mathrm{utt_0}$ ({$\mathrm{intent_0}$}) + $\mathrm{utt_1}$ ({$\mathrm{intent_1}$}) \\
        
        \bottomrule
    \end{tabular}
    
    \caption{Specification of the prompt used for the Generative Concatenation Approach with $N=3$.}
    \label{table:appendix_concatenation_prompt}
\end{table*}

\subsection{For Evaluating Multi-intent Detection}
\label{subsec:prompt_evaluate}
Table \ref{table:appendix_chatgpt_prompt} is the prompt for evaluating MID using ChatGPT in our study. Each dataset offers its distinct set of labels and examples to detect multiple intents of the given query.
In this study, we expanded the dataset up to 150 single intents into multi-label settings, significantly expanding the label space.
This expansion limited to the capacity to provide all examples for each prompt. 
The lack of a standardized approach to these few-shot prompts would be a notable challenge.

\begin{table*}
    \small
    \centering
    \renewcommand{\arraystretch}{1.5}
    \setlength{\tabcolsep}{0.5em}
    
    \begin{tabular}{>{\raggedright\arraybackslash}p{15cm}}
        \toprule
        You are an Intent Detection Model on single utterance. \\
        
        \textbf{[Task Definition]} Detect single or more intent(s) of each utterance, but you can only classify UP TO 3 most plausible intents on 1 utterance.\\
        
        \textbf{[Intents]} \texttt{atis\_airport, atis\_ground\_service, atis\_abbreviation, atis\_city, atis\_aircraft, atis\_ground\_fare, atis\_flight, ...} \\
        
        \textbf{[Answer format]} If more than one, concatenate with `\#', such as \{Intent\}\#\{Intent\}. 
        
        e.g. \texttt{atis\_ground\_fare\#atis\_distance} \\
        \arrayrulecolor{lightgray}\midrule\arrayrulecolor{black} 
        
            \textbf{[Example 1]}
            \begin{quote}
                {[Utterance]} does delta aircraft fly dc10
                
                {[Answer]} \texttt{atis\_aircraft} 
            \end{quote}
            
            \textbf{[Example 2]}
            \begin{quote}
                {[Utterance]} which airline has more business class flights than any other airline and what city is the airport mco in
                
                {[Answer]} \texttt{atis\_airline\#atis\_city} 
            \end{quote}
            
            \textbf{[Example 3]}
            \begin{quote}
            {[Utterance]} what does the fare code qx mean, what is the distance between pittsburgh airport and downtown pittsburgh and what is restriction ap80
            
            {[Answer]} \texttt{atis\_abbreviation\#atis\_distance\#atis\_restriction} 
            \end{quote}
            \\
            \arrayrulecolor{lightgray}\midrule\arrayrulecolor{black}
        
        \textbf{[Query]} Detect a single or up to 3 intent(s) on this following utterance. : $\mathrm{utt}$ \\
        
        \bottomrule
    \end{tabular}
    
    \caption{Specification of the prompt used for addressing MID within the in-context learning framework. (Dataset: MixATIS).}
    \centering
    \label{table:appendix_chatgpt_prompt}
\end{table*}

\end{document}